\documentclass[11pt, letterpaper]{article}

\usepackage{amsmath}
\usepackage{amssymb}
\usepackage{amsthm}
\usepackage{graphicx}
\usepackage{hyperref}
\usepackage{url}
\usepackage{algorithm}
\usepackage{algorithmic}
\usepackage{caption}
\usepackage{subcaption}
\usepackage{booktabs}
\usepackage{authblk}
\usepackage{appendix}
\usepackage[utf8]{inputenc}
\usepackage[T1]{fontenc}
\usepackage[english]{babel}
\usepackage{geometry}
\usepackage{xcolor}
\usepackage{fancyhdr}
\usepackage{microtype}
\usepackage{parskip}
\usepackage{tabularx}
\usepackage{enumitem}
\usepackage{siunitx}
\usepackage{adjustbox}

\hypersetup{
    colorlinks=true,
    linkcolor=blue,
    filecolor=magenta,      
    urlcolor=cyan,
    citecolor=red,
}

\definecolor{headercolor}{RGB}{0, 50, 100}
\newcommand*\samethanks[1][\value{footnote}]{\footnotemark[#1]}

\title{Radiology-Llama2: Best-in-Class Large Language Model for Radiology}

\author[1]{Zhengliang Liu \thanks{Co-first authors.}}
\author[1]{Yiwei Li \samethanks}
\author[1]{Peng Shu \samethanks}
\author[2]{Aoxiao Zhong}
\author[3]{Longtao Yang}
\author[3]{Chao Ju}
\author[1]{Zihao Wu}
\author[4]{Chong Ma}
\author[5]{Jie Luo}
\author[5]{Cheng Chen}
\author[5]{Sekeun Kim}
\author[5]{Jiang Hu}
\author[1]{Haixing Dai}
\author[1]{Lin Zhao}
\author[6]{Dajiang Zhu}
\author[3]{Jun Liu}
\author[7]{Wei Liu}
\author[8,9,10]{Dinggang Shen}
\author[1]{Tianming Liu}
\author[5]{Quanzheng Li}
\author[5]{Xiang Li}

\affil[1]{School of Computing, University of Georgia}
\affil[2]{Department of Electrical Engineering, Harvard University}
\affil[3]{Department of Radiology, Second Xiangya Hospital}
\affil[4]{School of Automation, Northwestern Polytechnical University}
\affil[5]{Department of Radiology, Massachusetts General Hospital and Harvard Medical School}
\affil[6]{Department of Computer Science and Engineering, University of Texas at Arlington}
\affil[7]{Department of Radiation Oncology, Mayo Clinic}
\affil[8]{School of Biomedical Engineering, ShanghaiTech University}
\affil[9]{Shanghai United Imaging Intelligence Co., Ltd.}
\affil[10]{Shanghai Clinical Research and Trial Center}

\date{}

\begin{document}

\maketitle

\begin{abstract}
This paper introduces Radiology-Llama2, a large language model specialized for radiology through a process known as instruction tuning. Radiology-Llama2 is based on the Llama2 architecture and further trained on a large dataset of radiology reports to generate coherent and clinically useful impressions from radiological findings. Quantitative evaluations using ROUGE metrics on the MIMIC-CXR and OpenI datasets demonstrate that Radiology-Llama2 achieves state-of-the-art performance compared to other generative language models, with a Rouge-1 score of 0.4834 on MIMIC-CXR and 0.4185 on OpenI. Additional assessments by radiology experts highlight the model's strengths in understandability, coherence, relevance, conciseness, and clinical utility. The work illustrates the potential of localized language models designed and tuned for specialized domains like radiology. When properly evaluated and deployed, such models can transform fields like radiology by automating rote tasks and enhancing human expertise. 
\end{abstract}

\section{Introduction}

Transformer-based large language models (LLMs) such as ChatGPT and GPT-4 have shown impressive capabilities in natural language processing \cite{wang2023review,liu2023evaluating,liu2023summary,holmes2023evaluating}. The development in transformer-based NLP models has also spurred advancements in developing and applying transformer-based models in computer vision \cite{dai2023samaug,zhang2023segment,li2023artificial,zhang2023beam} and other modalities \cite{bi2023community,zhong2023chatabl,zhang2023differentiating,ding2023deep,liu2023deid,dai2023chataug,liao2023mask,ding2023deep,ding2022accurate,liu2023context,liu2022discovering,dai2023samaug,dai2022graph}. Since November 2022, inspired by the versatile capabilities and wide popularity of ChatGPT, LLMs have been applied in clinical studies \cite{guan2023cohortgpt}, pharmacy \cite{liu2023pharmacygpt}, radiology \cite{liu2023radiology,wu2023exploring,ma2023impressiongpt,chang2023meta}, Alzheimer’s disease \cite{dai2023ad,cai2023exploring}, agriculture \cite{rezayi2023exploring} and brain science research \cite{zhao2023brain}. 

However, their application in specialized domains like healthcare has been limited.

First, localized large language models are a must for real world healthcare, since hospitals cannot share data or upload data to commercial models such as ChatGPT or GPT-4 due to privacy regulations \cite{liu2023deid,liao2023differentiate,liu2023summary}.

In addition, LLMs trained on general domain, such as ChatGPT \cite{openaiIntroducingChatGPT}, GPT-4 \cite{openai2023gpt} and PaLM 2 \cite{anil2023palm}, lack medical knowledge in specialized domains such as radiology, and it is necessary to design a model that is properly trained on domain data that is clinically meaningful.

Moreover, our Radiology-Llama2 perfectly imitates the style or radiologists, yet models like ChatGPT generate comprehensive but Wikipedia-like responses unlike the concise and simple language style of real radiologists that facilitates quick information exchange.
\begin{figure}
    \centering
    \includegraphics[scale=0.1]{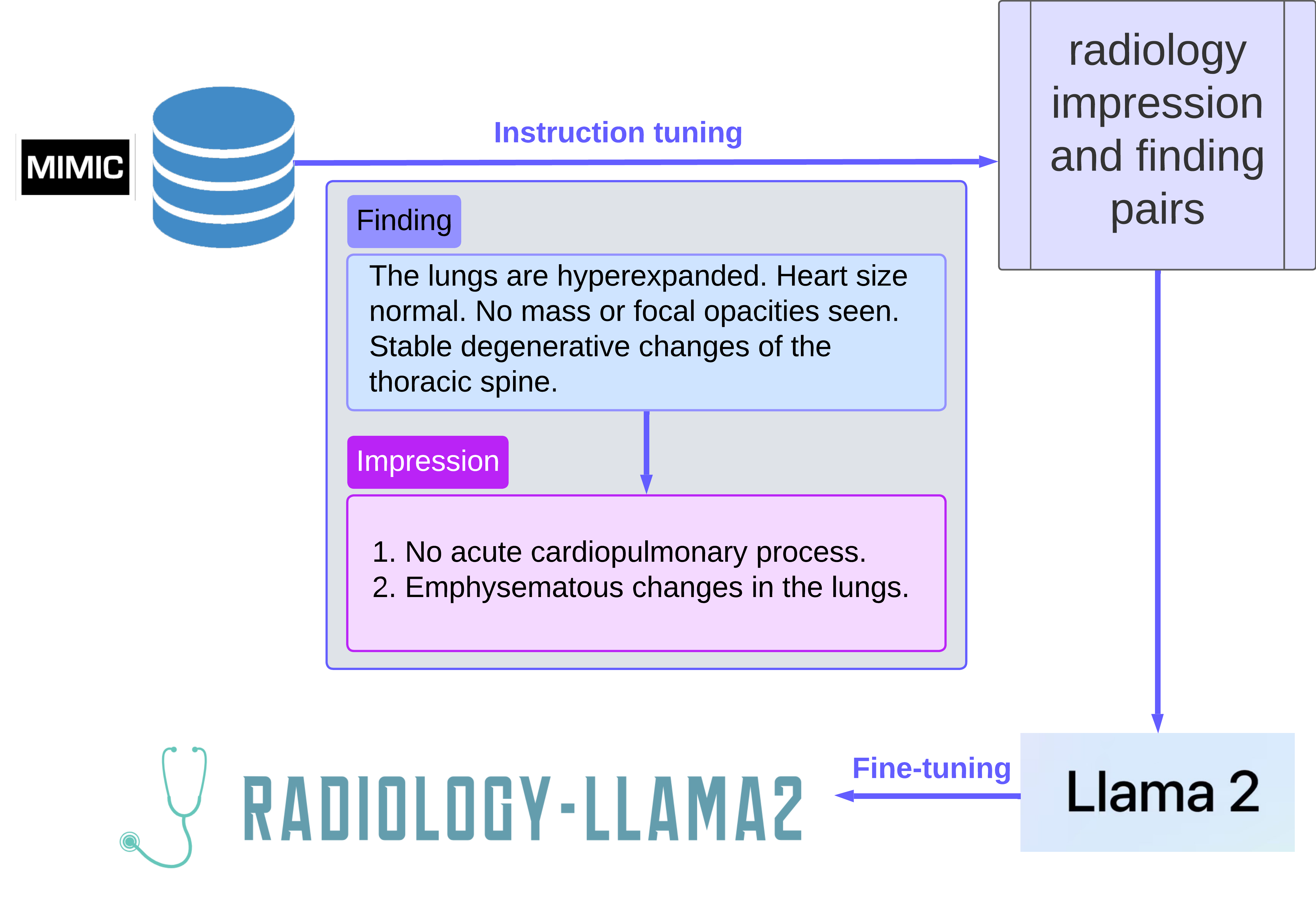}
    \caption{The overall framework of Radiology-Llama2.}
    \label{fig:framework}
\end{figure}
Finally, this work opens the door for personalized radiological assistants that are tailored to the style of individual physicians \cite{lecler2023revolutionizing}.

This work addresses this gap through Radiology-Llama2, an LLM tailored for radiology through instruction tuning to generate radiology impressions from findings. Evaluations show it surpasses general LLMs in coherence, conciseness and clinical utility of generated impressions.

\begin{itemize}
  \item \textbf{State-of-the-Art Performance:} Outperforms any other language models in deriving clinical impressions \cite{liu2023evaluating}, setting a new benchmark on MIMIC-CXR and OpenI datasets.
  
  \item \textbf{Flexibility and Dynamism:} Unlike its BERT-based counterparts \cite{devlin2018bert,zhou2023comprehensive}, Radiology-Llama2 is not tied to a specific input structure, allowing for a broader range of inputs and adaptability to different tasks within radiology, including complex reasoning.
  
  \item \textbf{Clinical Usability with Conversational Capabilities:} Generative LLMs offers inherent conversational functionality \cite{li2023artificial}, enabling it to provide contextual insights and responses in a human-like manner. This makes Radiology-Llama2 particularly useful for medical professionals in a clinical setting, enhancing both diagnosis and reporting.
\end{itemize}

\section{Related work}

\subsection{Large Language Models (LLMs)}
Recent developments in NLP are marked by the emergence of LLMs such as GPT-3 \cite{brown2020language}, GPT-4 \cite{openai2023gpt}, PaLM \cite{chowdhery2022palm}, and PaLM-2 \cite{anil2023palm}. Contrasting the earlier pre-training and fine-tuning approach observed in BERT \cite{devlin2018bert}, GPT \cite{radford2018improving}, GPT-2 \cite{radford2019language}, and their variants \cite{liu2019roberta,liao2023mask,zhou2023comprehensive}, these new LLMs exhibit few-shot and zero-shot learning capabilities using in-context learning. Furthermore, open-source models like LLaMA \cite{touvron2023llama} and Bloom \cite{scao2022bloom} have entered the scene, promoting broader accessibility. 

There's also an increasing interest in instruction-tuned models such as Alpaca \cite{stanfordStanfordCRFM}, StableLM \cite{stabilityStabilityLaunches}, and Dolly \cite{databricksFreeDolly}. 

\subsection{Domain-Specific Language Models (DSLMs)}

DSLMs, such as AgriBERT \cite{rezayi2022agribert}, are tailored to specific domains, aimed at optimal performance in related tasks. Specifically, AgriBERT is trained on agricultural texts, making it suitable for tasks in agriculture. SciEdBERT \cite{liu2023context} is designed for the educational sector and focuses on middle school chemistry and physics, offering insights into evaluating students' responses. ClinicalRadioBERT \cite{rezayi2022clinicalradiobert}, in the healthcare sector, is adept at radiation oncology and emphasizes its training from clinical notes and related literature. These DSLMs highlight the potential and adaptability of specialized models across various sectors \cite{liu2023context,cai2022coarse,zhou2023fine,liao2023mask}. 

\section{Methodology}
Radiology-Llama2 is trained on a large radiology dataset using instruction tuning to generate radiology impressions from findings. Evaluations by radiologists show it surpasses general LLMs in coherence, conciseness and clinical utility of generated impressions.
\begin{figure}
    \centering
    \includegraphics[scale=0.5]{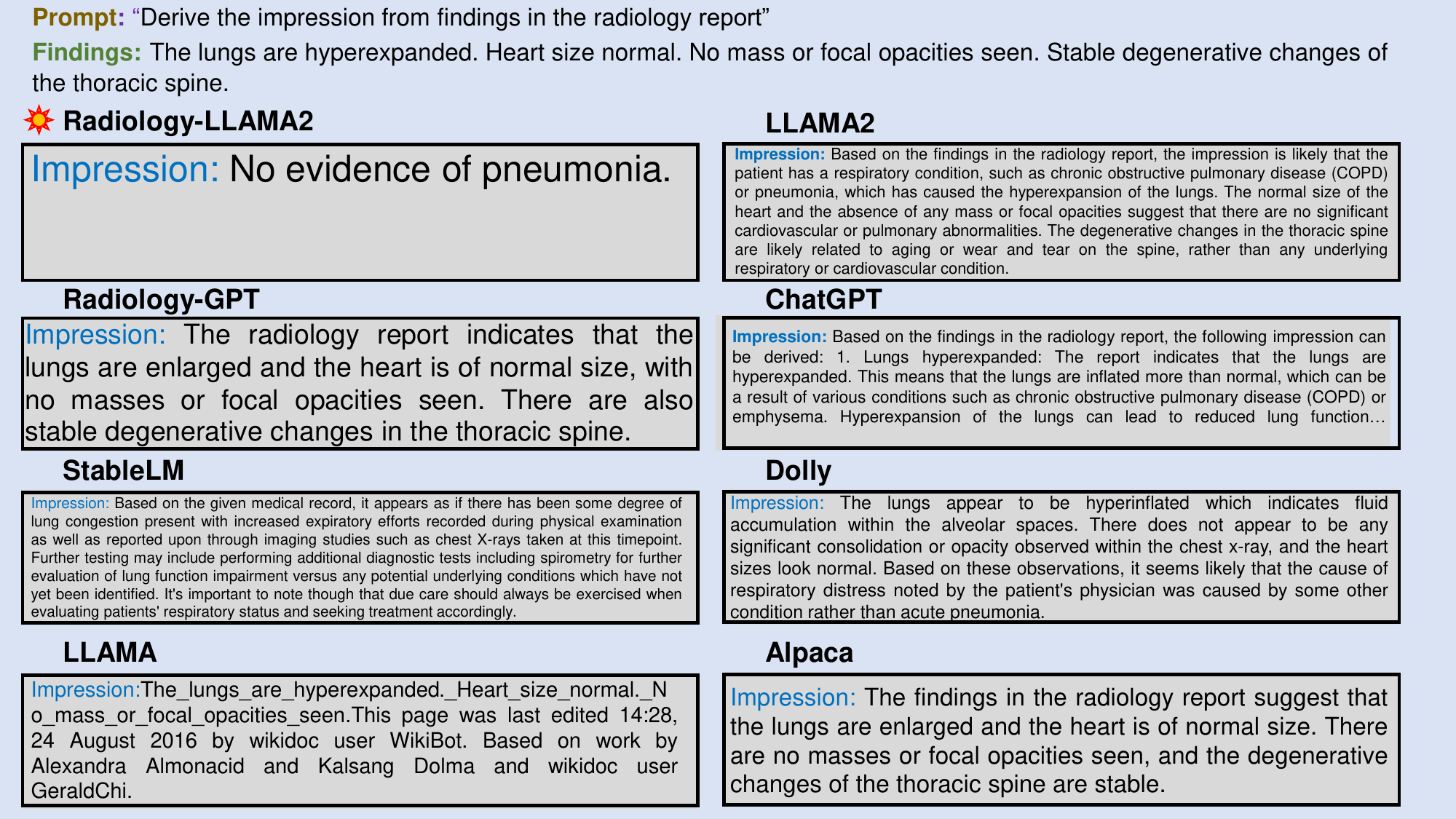}
    \caption{Performance of different LLMs on the Radiology task example.}
    \label{fig:exp1}
\end{figure}
\subsection{Datasets}
\subsubsection{MIMIC-CXR Dataset}
MIMIC-CXR is a large chest radiographs dataset which consists of 227,835 imaging studying based on 65,379 patients presenting to the Beth Israel Deaconess Medical Center Emergency Department between 2011–2016 \cite{johnson2019mimic}. There are 377,110 available images in the dataset where each imaging studying contains one or more images (typically a frontal view and a lateral view). This dataset also has the corresonding free-text radiology reports and has been de-identified to ensure the US Health Insurance Portability and Accountability Act of 1996 (HIPAA) Safe Harbor requirements. Widely application of MIMIC-CXR dataset has been implemented in computer vision, natural language processing and decision support etc.

\subsubsection{OpenI Dataset}
OpenI dataset is a publicly available dataset aiming to make clinical documents for secondary use in the region of research and education \cite{demner2016preparing}. This dataset collects 8121 images form the hospitals' picture archiving systems accompanied with 3996 corresponding radiology reports from the Indiana Network. Manual coding has been added into the radiologist reports in order to increase the relevancy of the retrieved clinical documents. Similar to MIMIC-CXR dataset, OpenI uses manually verification after the automatic method to achieve de-identification.

\subsection{Instruction Tuning}

\subsubsection{Principle and Objective}

Instruction tuning \cite{wei2021finetuned,ouyang2022training} is a foundational component of the Radiology-Llama2 framework. Instruction tuning addresses the fundamental disconnect between the traditional training objectives of LLMs and the user-specific goals of instruction following. This technique involves additional training using pairs of human-specified instructions and corresponding desired outputs. It serves to align the model with task-specific user objectives, enhance model controllability, and allow for rapid domain-specific adaptation, all while maintaining computational efficiency.

To bolster learning, instructions are formatted specifically. For instance, the "Findings" text is supplied with a succinct instruction, such as "Derive the impression from findings in the radiology report", while the "Impression" text from the same report serves as the target output. This approach calibrates the model in alignment with the desired task, thus yielding an instruction-adhering language model optimized for radiology reports.

\subsubsection{Domain-Specific Knowledge Acquisition}
Through training on domain-specific data, the model is also adept at assimilating domain-specific knowledge that is quintessential to radiology. Consequently, Radiology-Llama2 is proficient in capturing language patterns, terminologies, and logical reasoning essential for interpreting radiology reports.

The initial instruction tuning is centered on the "Findings" to "Impression" conversion, which holds significant clinical value. Recognizing the potential of diverse instruction pairs in radiology, ongoing engagements with radiologists are aimed at formulating a diverse set of clinically pertinent instruction tuning pairs to augment the capabilities of Radiology-Llama2.

\subsection{Experimental Setting}

The Radiology-Llama2 model employs an advanced training regimen. For reference, when training Radiology-GPT with the Llama2-7b-chat \cite{touvron2023llama} base model, the training was facilitated by Low-Rank Approximations (LoRA) \cite{hu2021lora}. The choice of LoRA was motivated by its compact size and portability which are conducive to model sharing and deployment.

The experimental setting for the training comprised the following configurations:
\begin{itemize}
    \item Batch size: 128
    \item Learning rate: $3 \times 10^{-4}$
    \item Weight decay: 0.01
\end{itemize}

\begin{flushleft}
LoRA parameters:
\begin{itemize}
    \item \texttt{lora\_r} (rank of low-rank factorization): 8
    \item \texttt{lora\_alpha} (scaling factor for the rank): 16
    \item \texttt{lora\_dropout}: 0.05
\end{itemize}
\end{flushleft}

The target modules for LoRA were set to \texttt{"q\_proj"} and \texttt{"v\_proj"}, corresponding to the query and value matrices in the self-attention mechanism of the transformer architecture. The training was conducted on a server equipped with 4 Nvidia A100 80GB GPUs.

\section{Results}
The present study evaluated the performance of various large language models on two key datasets pertinent to radiology: MIMIC-CXR and OpenI. The assessment employed Rouge-1, Rouge-2, and Rouge-L as the primary metrics, given their widespread acceptance for evaluating the quality of generated text.
\begin{figure}
    \centering
    \includegraphics[scale=0.5]{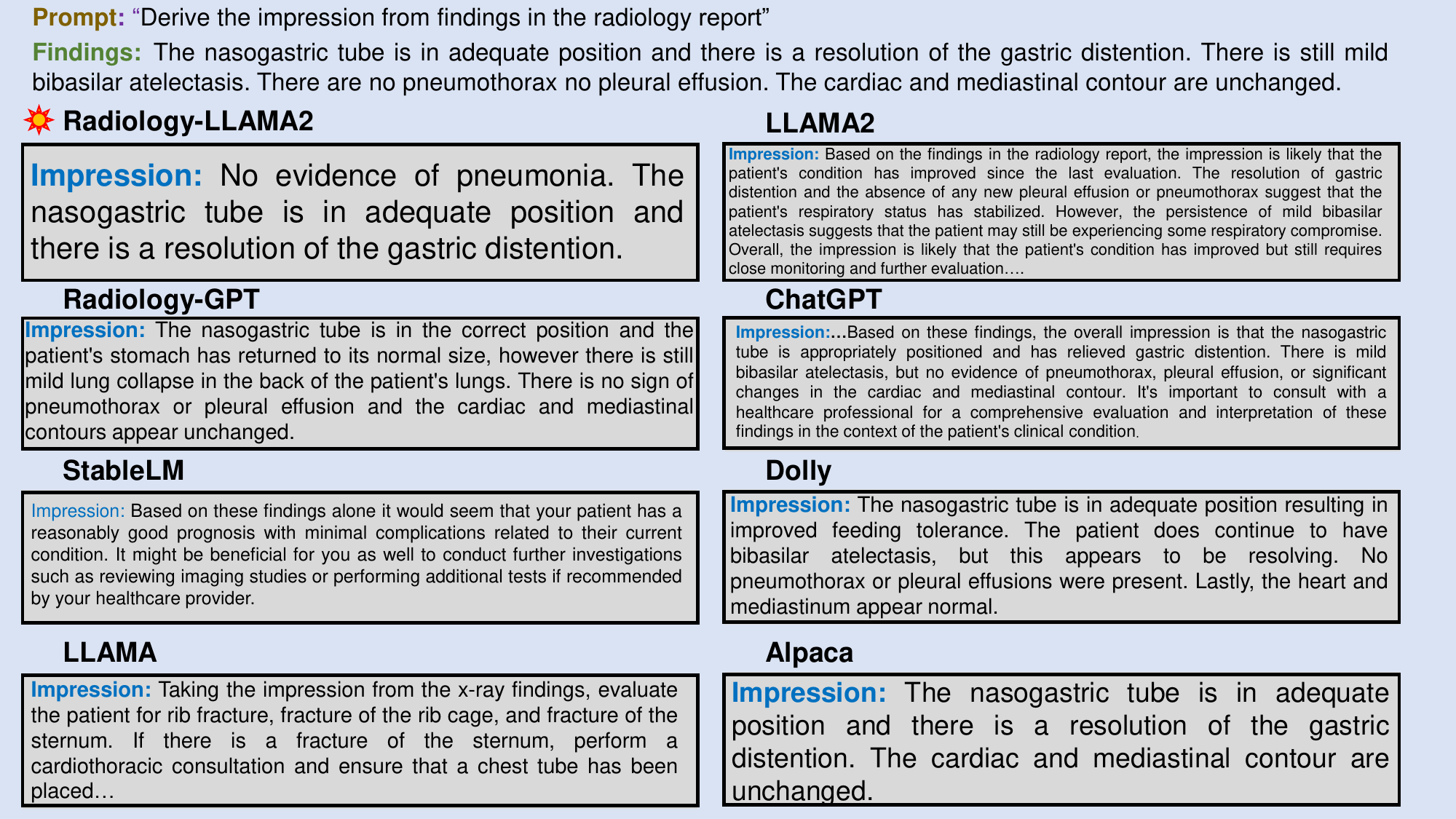}
    \caption{Performance of different LLMs on the Radiology task example.}
    \label{fig:exp2}
\end{figure}
\subsection{Quantitative Assessment}
Radiology-Llama2 significantly outperforms all comparison models across all ROUGE \cite{lin2004rouge} metrics: ROUGE-1, ROUGE-2, and ROUGE-L, on both MIMIC-CXR and OpenI datasets. Results can seen in Table \ref{tab:rouge_scores_10} and Table \ref{tab:rouge_scores_all}.

For the MIMIC-CXR dataset, Radiology-Llama2 achieves scores of 0.4834 in ROUGE-1, 0.324 in ROUGE-2, and 0.4427 in ROUGE-L. These scores are markedly higher than those of the second-best performing model, Anthropic Claude2, which manages 0.3177 in ROUGE-1 and 0.153 in ROUGE-2. This demonstrates that Radiology-Llama2 not only captures a higher proportion of overlapping unigrams between the generated and reference summaries but also maintains this superiority in capturing bigrams and maintaining a longer sequence of content overlap.

Similarly, in the OpenI dataset, Radiology-Llama2 sustains its exemplary performance, recording scores of 0.4185 in ROUGE-1, 0.2569 in ROUGE-2, and 0.4087 in ROUGE-L. In comparison, the second-best model, Anthropic Claude2, scores 0.2372 in ROUGE-1 and 0.1259 in ROUGE-2. The substantial gap between the two models across all metrics underscores Radiology-Llama2's robustness and generalizability across datasets.

On the lower end of the performance spectrum, Baichuan-7B exhibits exceptionally low scores on both datasets. For instance, its ROUGE-2 score is a meager 0.0057 on the MIMIC-CXR dataset, and its ROUGE-L score is only 0.0029. This emphasizes the limitations of such models in capturing even the basic elements of content overlap.
\begin{figure}
    \centering
    \includegraphics[scale=0.35]{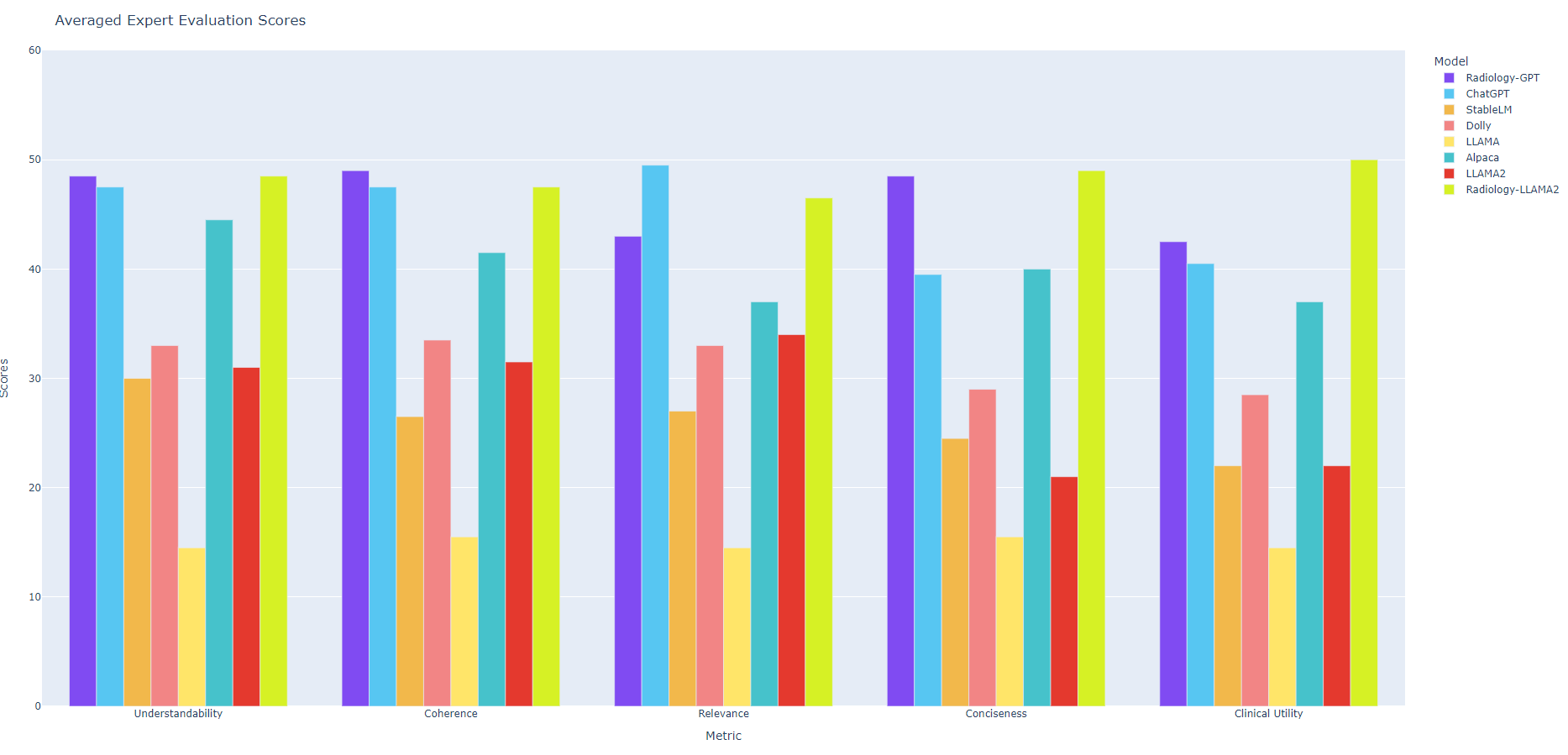}
    \caption{ Radiologists' evaluation on the diagnostic results of different LLMs based on radiology reports.}
    \label{fig:evaluation}
\end{figure}
It is also noteworthy that Radiology-Llama2 maintains its superiority not only in terms of single-term overlap but also in capturing a longer chain of content, as reflected in its ROUGE-L scores.

This paper has provided two specific examples. The first example can be seen in Figure \ref{fig:exp1} and Figure \ref{fig:exp2}, the finding is “The lungs are hyperexpanded. Heart size normal. No mass or focal opacities seen. Stable degenerative changes of the thoracic spine.” Six LLMs are required to derive the impression from the findings in the radiology report. It can be seen that LLAMA2, ChatGPT and Alpaca all understand the content of the radiology report, however, their answers are too trivial and fail to catch the most important points, which will make their given impression hard to understand and increase the difficulty to be used clinically. In contrast, although StableLM, Dolly and LLAMA basically understand the intention of the text, their answers are not satisfactory. Their answers are not only redundant but also contain a lot of irrelevant information. Radiology-GPT, which has been performed a fine-tuning on a specific data set, has the answer which is obviously more capable and concise, but the gap with the impression given by Radiologist is not small. Therefore, in comparison, it can be clearly seen that the excellent performance of Radiology-LLAMA2, the answer of Radiology-LLAMA2 is not only accurate and concise, but also has the report style which is closest to Radiologist. This intuitively proves the performance of Radiology-LLAMA2 and proves its strong clinical application potential.

In summary, Radiology-Llama2 consistently demonstrates superior performance in generating clinically relevant and coherent radiology reports, as evidenced by its high ROUGE scores across multiple datasets.

\begin{table}[ht]
\centering
\begin{tabularx}{\textwidth}{l *{8}{X}}
\toprule
         & Llama2 & Radiology-Llama2 & Radiology-GPT & ChatGPT & StableLM & Dolly & Llama & Alpaca \\
\midrule
rouge-1   & 0.201  & \textbf{0.4726}           & 0.3123        & 0.0774  & 0.0734   & 0.1866 & 0.0873 & 0.3464 \\
rouge-2   & 0.0969 & \textbf{0.2948}           & 0.1758        & 0.0432  & 0.0085   & 0.0717 & 0.0426 & 0.2071 \\
rouge-L   & 0.1578 & \textbf{0.3757}           & 0.2518        & 0.0702  & 0.0516   & 0.1572 & 0.0776 & 0.2935 \\
\bottomrule
\end{tabularx}
\caption{Rouge scores of Radiologist's evaluations on 10 MIMIC-CXR-OpenI examples.}
\label{tab:rouge_scores_10}
\end{table}

\begin{table}[ht]
\centering
\sisetup{table-format=1.4, table-number-alignment=center}
\renewcommand{\arraystretch}{1.2} % Adds padding to cells
\begin{tabular*}{\textwidth}{
  @{\extracolsep{\fill}}
  l
  *{3}{S}
  *{3}{S}
  @{}
}
\toprule
& \multicolumn{3}{c}{MIMIC-CXR} & \multicolumn{3}{c}{OpenI} \\
\cmidrule(lr){2-4} \cmidrule(lr){5-7}
& {Rouge-1} & {Rouge-2} & {Rouge-L} & {Rouge-1} & {Rouge-2} & {Rouge-L} \\
\midrule
Radiology-Llama2   & \textbf{0.4834} & \textbf{0.324} & \textbf{0.4427} & \textbf{0.4185}  & \textbf{0.2569} & \textbf{0.4087} \\
Llama2~\cite{touvron2023llama}   & 0.1726 & 0.0711 & 0.1322 & 0.0848 & 0.0205 & 0.0712 \\
Ziya-LLaMA-13B-v1~\cite{fengshenbang}  &0.2121&0.0968&0.17&0.1101&0.0316&0.0926\\
ChatGPT~\cite{openaiIntroducingChatGPT} &0.2048&0.0996&0.1702&0.1203&0.037&0.1052 \\
GPT-4~\cite{openai2023gpt}  &0.1995&0.0858&0.1575&0.1171&0.0343&0.0975 \\
ChatGLM2-6B~\cite{du2022glm} &0.2042&0.0964&0.1605&0.1094&0.0331&0.0909\\
MOSS-MOON-0030sft~\cite{sun2023moss} &0.2203&0.0914&0.1757&0.1402&0.0341&0.1241\\
PaLM2~\cite{anil2023palm} &0.2749&0.1442&0.2281&0.1386&0.0477&0.1194\\
Sensenova~\cite{Sensenova} & 0.1136 & 0.0304 & 0.0809 & 0.0634 & 0.0106 & 0.051\\
Anthropic Claude2~\cite{Claude2} & 0.3177 & 0.153 & 0.256 & 0.2372 & 0.1259 & 0.2193\\
BayLing-7B~\cite{bayling} &0.2149&0.107&0.1747&0.1252&0.0389&0.1044\\
XrayGLM~\cite{wang2023XrayGLM} & 0.1104 & 0.0468 & 0.1211 & 0.0622 & 0.0183 & 0.0599\\
Bianque v2~\cite{chen2023bianque1} & 0.0294 & 0.0072 & 0.0281 & 0.0227 & 0.0036 & 0.0222\\
XrayPULSE~\cite{xraypulse} & 0.0592 & 0.0061 & 0.0398 & 0.0293 & 0.0011 & 0.0239 \\
DoctorGLM~\cite{xiong2023doctorglm} & 0.1853 & 0.0916 & 0.153 & 0.0996 & 0.0329 & 0.0861\\
BenTsao~\cite{wang2023huatuo} & 0.1319 & 0.0618 & 0.1126 & 0.0804 & 0.0368 & 0.0711\\
Baichuan-7B~\cite{baichuan} & 0.2379 & 0.0057 & 0.0029 & 0.003 & 0.0009 & 0.0028\\
\bottomrule
\end{tabular*}
\caption{Rouge scores on MIMIC-CXR and OpenI datasets.}
\label{tab:rouge_scores_all}
\end{table}

\subsection{Expert Evaluation}
To supplement the quantitative evaluations, we conducted an expert-based assessment of the models. We randomly selected 10 records each from MIMIC-CXR and OpenI datasets and had two experienced radiologists manually evaluate the generated radiology impressions based on five key criteria: Understandability, Coherence, Relevance, Conciseness, and Clinical Utility.

\begin{itemize}
    \item \textbf{Understandability:} Radiology-Llama2 and Radiology-GPT both stood out with a score of 48.5, suggesting that their generated impressions are highly understandable to radiologists. ChatGPT closely followed with a score of 47. On the other end of the spectrum, Llama scored the lowest at 14, indicating significant limitations in generating understandable content.

    \item \textbf{Coherence:} Radiology-Llama2 once again led the cohort with a score of 47.5, closely followed by Radiology-GPT at 48 and ChatGPT at 46.5. Llama lagged considerably, scoring only 13.5, which raises questions about its ability to produce coherent clinical text.

    \item \textbf{Relevance:} In terms of the relevance of the generated text to the radiology findings, ChatGPT surprisingly took the lead with a score of 49.5. Radiology-Llama2 was close behind with a 46.5, while Radiology-GPT scored 43.5. Llama and StableLM were the least relevant models, scoring 14 and 25, respectively.

    \item \textbf{Conciseness:} Radiology-Llama2 excelled in generating concise impressions, scoring the highest at 49. Llama, StableLM, and Dolly demonstrated shortcomings in this regard, scoring 14, 24.5, and 27.5 respectively.

    \item \textbf{Clinical Utility:} Radiology-Llama2 emerged as the most clinically useful model, garnering a top score of 50. In contrast, Llama scored the lowest in this metric as well, with a score of 13, highlighting its limited utility for clinical applications.
\end{itemize}

In addition, we evaluated the Rouge scores of a few select models on these 10 samples to cross-verify with the experts' evaluation. Please see \ref{tab:rouge_scores_10} for more details. 

Overall, Radiology-Llama2 consistently demonstrated superior performance across all five criteria, affirming its status as a highly effective tool for generating radiology impressions. While other models like Radiology-GPT and ChatGPT showed competence in certain areas, they were unable to match the all-around excellence of Radiology-Llama2. Models like Llama and StableLM displayed significant limitations, performing poorly across multiple criteria.

\section{Discussion}

\subsection{Application in Clinical Radiology Diagnosis}
The high Rouge scores attained by Radiology-LLAMA2 suggest that this language model has the potential for significant impact in clinical radiology diagnosis. Its ability to swiftly generate coherent and clinically relevant reports could be transformative, particularly in busy radiology departments where timely and accurate reporting is of the essence. By automating certain aspects of the reporting process, the model serves as a valuable assistive tool for radiologists, allowing them to focus more on complex cases that require nuanced human expertise.

\subsection{Need for Diverse Training Data}
While Radiology-LLAMA2 performs impressively on generating findings and impressions, its utility could be further broadened by incorporating more diverse forms of training data. For instance, it could be trained on instructional text for medical procedures, summaries of patient histories, or physician’s notes on differential diagnoses. By diversifying the data sources, the model would be better equipped to assist in various facets of radiological practice, such as recommending further diagnostic tests or suggesting possible treatment paths.

\subsection{Multimodality: Adding Image Capabilities}
The addition of image analysis capabilities would elevate Radiology-LLAMA2 from a text-generation model to a truly multimodal diagnostic tool. Future iterations could integrate machine learning algorithms for image recognition, enabling the model to make direct observations from X-rays, MRIs, or CT scans. This would potentially create a more holistic diagnostic process where textual and visual data are analyzed in tandem for more accurate and comprehensive diagnoses.

\subsection{Conversational Assistant to Radiologists}
Beyond report generation, Radiology-LLAMA2 could be developed into a conversational assistant that helps radiologists in real-time. This would enable a more dynamic interaction, where the model could assist in tasks ranging from quick data retrieval to offering second opinions on diagnoses. Such a system could act as a "second pair of eyes," providing immediate feedback and thus serving as a valuable safeguard against diagnostic errors.

\section{Conclusion}
Radiology-Llama2 demonstrates localized LLMs can transform radiology when designed appropriately. With proper oversight, it has great potential for clinical decision support and other applications. This work pave the way for specialized LLMs in other medical domains.

In conclusion, Radiology-Llama2 represents an important advance in applying LLMs to healthcare. With continued research into model design and evaluation, such specialized LLMs can enable breakthroughs in medical AI.

\bibliography{LLM_refs}
\bibliographystyle{unsrt}

\end{document}